\pdfoutput=1
\documentclass[conference]{IEEEtran}
\IEEEoverridecommandlockouts
\usepackage{cite}
\usepackage{amsmath,amssymb,amsfonts}
\usepackage{algorithmic}
\usepackage{graphicx}
\usepackage{textcomp}
\usepackage{xcolor}
\usepackage{booktabs}

\usepackage{tikz}
\usetikzlibrary{calc, patterns}
\usepackage{pgfplots}
\pgfplotsset{compat=1.15}
\definecolor{citeblue}{rgb}{0.1,0,.4}
\usepackage[pdftex%
,colorlinks=true%
,bookmarks=true%
,linkcolor=citeblue%
,citecolor=citeblue%
,urlcolor=blue%
,plainpages=false]{hyperref}
\usepackage{graphicx}
\usepackage{amsthm}
\newtheorem{definition}{Definition}

\DeclareMathOperator*{\naturals}{\mathbb{N}}
\newcommand{\terms}{\mathcal{T}}

\usepackage[linesnumbered,ruled,vlined]{algorithm2e}
\DontPrintSemicolon%
\SetKwData{false}{false}
\SetKwData{true}{true}
\SetKwInOut{Input}{input}
\SetKwInOut{Output}{output}
\SetKw{Break}{break}
\SetKw{Yield}{yield}
\SetKwRepeat{Do}{do}{while}

\def\miko{Mikol\'a\v s Janota}
\begin{document}

\newcommand\TheTitle{Fair and Adventurous Enumeration\\ of Quantifier Instantiations}
\title{\TheTitle*
\thanks{The results were supported by the Ministry of Education, Youth and Sports within the dedicated program ERC CZ under the project POSTMAN no.~LL1902.
This scientific article is part of the RICAIP project that has received funding from the European Union's Horizon 2020 research and innovation programme under grant agreement No 857306.}
}

\author{\IEEEauthorblockN{\miko}
\IEEEauthorblockA{\textit{Czech Technical University in Prague}\\
Prague, Czech Republic\\
0000-0003-3487-784X}
\and
\IEEEauthorblockN{Haniel Barbosa}
\IEEEauthorblockA{\textit{Universidade Federal de Minas Gerais} \\
Belo Horizonte, Brazil \\
000-0003-0188-2300}
\and
\IEEEauthorblockN{Pascal Fontaine}
\IEEEauthorblockA{\textit{University of Liège}\\
Liège, Belgium \\
0000-0003-4700-6031}
\and
\IEEEauthorblockN{Andrew Reynolds}
\IEEEauthorblockA{\textit{University of Iowa} \\
 USA\\
0000-0002-3529-8682}
}

\maketitle

\begin{abstract}
  SMT solvers generally tackle quantifiers by instantiating their variables with
tuples of terms from the ground part of the formula.
Recent enumerative approaches for quantifier instantiation
consider tuples of terms in some heuristic order.
%
%
%
This paper studies different strategies to order such tuples and their impact
on performance.
We decouple the ordering problem into two parts.
First is the order of the
sequence of terms to consider for each quantified variable, and
%
%
second is the order of the instantiation tuples themselves.
%
%
While the most and least preferred tuples, i.e. those with all variables
assigned to the most or least preferred terms, are clear, the combinations in
between allow flexibility in an implementation.
%
%
We look at principled strategies of complete enumeration, where some strategies
are more fair, meaning they treat all the variables the same but some strategies
may be more adventurous, meaning that they may venture further down the
preference list.
We further describe new techniques for discarding irrelevant instantiations
which are crucial for the performance of these strategies in practice.
These strategies are implemented in the SMT solver cvc5, where they contribute
to the diversification of the solver's configuration space, as shown by our
experimental results.


\end{abstract}



\section{Introduction}

While SMT (satisfiability modulo theory) solvers~\cite{Barrett2018} are used
successfully as decision procedures to automatically discharge quantifiers-free
proof obligations for many applications, there is an increasing need for tools
that can furthermore handle quantifiers.  Quantified languages however are most
often undecidable, or have prohibiting complexity.  Quantifier handling within
SMT solving is thus a challenge, and requires good heuristics.

Quantifier reasoning in SMT builds on the strength of SMT solvers, that is, their
ability to efficiently reason on ground formulas, and relies on
instantiation: ground consequences of quantified formulas are generated, and the
ground reasoner's view of the problem is gradually refined with these instances,
to embed knowledge from the quantified formula into ground reasoning.
The terms to generate instances may be generated using mostly syntactic methods,
e.g., E-matching~\cite{DBLP:journals/jacm/DetlefsNS05}, or semantic techniques
like model-based quantifier instantiation~\cite{ge-moura-cav09}.
But plain enumeration, done in a principled manner, can give surprisingly good
results, particularly in combination with other instantiation
techniques~\cite{DBLP:conf/tacas/ReynoldsBF18}.

A crucial aspect, when using enumeration-based instantiation, is to prioritize
the numerous, often infinite, potential instantiations.  When instantiating just
one variable, this is essentially a matter of prioritizing smaller terms that
are already present in the original formula, according to some order.
Quantified assertions however most often have many quantified variables, and
there is a lot of freedom on the order on tuples of terms to instantiate those.
We here investigate a few strategies based on different tuple orders, some
favoring fairness, some being more adventurous, and show that they are valuable
in a portfolio of enumerative instantiation strategies.  In \autoref{sec:masking}, we also present an elimination technique for redundant instantiations that significantly contributes to the improvement of enumeration-based instantiation.



\section{Background}\label{sec:preliminaries}

\begin{figure}[t]

  \centering
  \begin{tikzpicture}[scale=0.8, every node/.style={scale=0.7}]

  \node[left] at (-2,1.05) {Input};
  \node[left] at (-2,.65) {formula};

  \draw[->] (-3,.85) -- (-2,.85);

  \draw (-2,-.3) rectangle (4,2);

  \node[below right] at (-2,2) {SMT solver};

  \node at (1,.35) {Ground solver};
  \draw (0,0) rectangle (2,.7);

  \node at (1,1.35) {Instantiation};
  \draw (0,1) rectangle (2,1.7);

  \draw[->,>=stealth] (0,.35) .. controls (-.5,0.6) and (-.5,1.1) .. (0,1.35);

  \node[left] at (-.4,1) {Ground};
  \node[left] at (-.4,0.7) {assignment};

  \draw[<-,>=stealth] (2,.35) .. controls (2.5,0.6) and (2.5,1.1) .. (2,1.35);

  \node[right] at (2.4,.85) {Instances};

  \draw[->,>=stealth] (4,.85) .. controls (4.3,.9) .. (4.5,1);
  \draw[->,>=stealth] (4,.85) .. controls (4.3,.8) .. (4.5,.7);

  \node[right] at (4.5,1) {satisfiable};
  \node[right] at (4.5,.7) {unsatisfiable};

  \node at (4.8,.3) {Or infinite};
  \node at (4.8,0) {loop};

  \end{tikzpicture}

  \caption{The SMT instantiation loop.}
  \label{fig:instloop}

\end{figure}
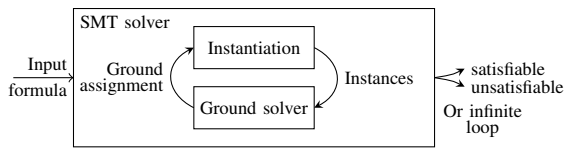

Originally, SMT solvers were essentially decision procedures for ground (i.e., quantifier free) problems in a combination of decidable languages, containing e.g., operators to handle arrays, linear arithmetic expressions, bitvectors, and uninterpreted predicates and functions.  They excel at deciding the satisfiability of large formulas in these languages.  As a toy example, consider the (satisfiable) conjunctive set of formulas
$$
\{R(a), \neg S(b), a = b\}.
$$
It belongs to the quantifier-free fragment of first-order logic, and as such, is decided by many SMT solvers.  Quantifier reasoning in modern SMT solvers builds on this.  The input formula, possibly after a pre-processing phase, is first given to the ground solver.  From the point of view of this ground solver, each quantified formula is abstracted into a distinct propositional variable.  As an example, the conjunctive set
$$
\{R(a), \neg S(b), a = b, \forall x\,.\, R(x) \Rightarrow S(x)\}
$$
is understood by the ground solver as the previous ground set, augmented with an abstract proposition $Q$ corresponding to $\forall x\,.\, R(x) \Rightarrow S(x)$.  
Then the ground solver provides a satisfying assignment for the ground part of the formula, including a valuation of the propositional variables abstracting the quantified formulas (in our case $Q$ must be true).  The instantiation module recovers the quantified formulas associated to these variables, and generates new instances of the quantified formulas to the ground reasoner (\autoref{fig:instloop}).  In our toy example such an instance could be
$$
Q \Rightarrow \left( R(a) \Rightarrow S(a) \right),
$$
which would render the problem unsatisfiable at the ground level.  In general,
the instantiation loop is iterated until the ground reasoner is able to conclude that the formula is unsatisfiable, a time out is reached, or no instance can be deduced anymore.  
In this paper, we focus on refutations only and will not consider the last case.

Thanks to the Herbrand Theorem (see e.g.,~\cite{DBLP:conf/tacas/ReynoldsBF18}),
with fair enumeration of instances using all possible terms built on the
appropriate set of symbols, SMT solving is refutationally complete for
satisfiability modulo well-behaved first-order theories.  Since typical SMT
inputs contain hundreds of quantified formulas with many nested quantifiers, on
a language with often infinitely many terms, the number of possible instances is
very large, and most often infinite.  It is crucial to quickly find out the
right instances, otherwise the ground solver will be overwhelmed by the amount
of instances.  For a quantified formula $\forall x_1\dots x_n\,.\,\varphi$ with
$n$ variables, this boils down to order $n$-tuples of ground terms to prioritize
instantiation.


\section{Enumeration Strategies}%
\label{sec:enumeration_strategies}
We start by the assumption that for each variable $x_i$ there is a sequence of
terms
$\terms_i=t_i^1,t_i^2,\dots$,
which are the possible candidates for instantiation into the
variable~$x_i$.  We further assume that this sequence of terms is sorted by
some given preference, i.e., that $t_i^j$ is  more likely to  yield a useful
instantiation than the candidate $t_i^{j'}$ with $j<j'$.  This lets us focus on
the indices into the sequences of terms, rather than on the terms themselves.
An instantiation, i.e., a tuple of terms, is uniquely represented as an
$n$-tuple of indices.

While this setup already assumes a given order on the terms for the individual
variables, it does not tell us how to order the actual tuples.  Clearly,
the tuple of indices $(0,\dots, 0)$ is the most advantageous and
$(|\terms_1| - 1,\dots, |\terms_n| - 1)$ is the least advantageous one.
However, it is unclear whether $(0, 1, 1)$ is more advantageous than $(0, 0,
2)$, or the other way around.  This motivates   our quest for different enumeration  strategies.  A general notion from multi-objective optimization is useful:   \emph{Pareto-optimal}
solutions are such that improving any criterion worsens some other.

\begin{figure}[th]
  \centering
\begin{tikzpicture}[xscale=0.9,yscale=0.67, every node/.style={scale=0.67}]
  \node (000) at (10,10) {$000$};

  \node (100) at  (9,9) {$100$};
  \node (010) at (10,9) {$010$};
  \node (001) at (11,9) {$001$};
  \draw[color=gray] (000) -- (100);
  \draw[color=gray] (000) -- (010);
  \draw[color=gray] (000) -- (001);

  \node (200) at  (8,8) {$200$};
  \node (110) at (8.8,8) {$110$};
  \node (101) at (9.6,8) {$101$};
  \node (020) at (10.4,8) {$020$};
  \node (011) at (11.2,8) {$011$};
  \node (002) at (12,8) {$002$};
  \draw[color=gray] (100) -- (200);
  \draw[color=gray] (100) -- (110);
  \draw[color=gray] (100) -- (101);

  \draw[color=gray] (010) -- (110);
  \draw[color=gray] (010) -- (020);
  \draw[color=gray] (010) -- (011);

  \draw[color=gray] (001) -- (101);
  \draw[color=gray] (001) -- (011);
  \draw[color=gray] (001) -- (002);

  \node (300) at  (6.4,7) {$300$};
  \node (210) at  (7.2,7) {$210$};
  \node (201) at  (8,7) {$201$};
  \node (120) at  (8.8,7) {$120$};
  \node (111) at  (9.6,7) {$111$};
  \node (030) at  (10.4,7) {$030$};
  \node (021) at  (11.2,7) {$021$};
  \node (102) at  (12,7) {$102$};
  \node (012) at  (12.8,7) {$012$};
  \node (003) at  (13.6,7) {$003$};

  \draw[color=gray] (200) -- (300);
  \draw[color=gray] (200) -- (210);
  \draw[color=gray] (200) -- (201);

  \draw[color=gray] (110) -- (210);
  \draw[color=gray] (110) -- (120);
  \draw[color=gray] (110) -- (111);

  \draw[color=gray] (101) -- (201);
  \draw[color=gray] (101) -- (111);
  \draw[color=gray] (101) -- (102);

  \draw[color=gray] (020) -- (120);
  \draw[color=gray] (020) -- (030);
  \draw[color=gray] (020) -- (021);

  \draw[color=gray] (011) -- (111);
  \draw[color=gray] (011) -- (021);
  \draw[color=gray] (011) -- (012);

  \draw[color=gray] (002) -- (102);
  \draw[color=gray] (002) -- (012);
  \draw[color=gray] (002) -- (003);

\end{tikzpicture}

  \caption{Pareto graph for 3 variables  with 4 candidate terms for each.}%
  \label{fig:graph}

\end{figure}
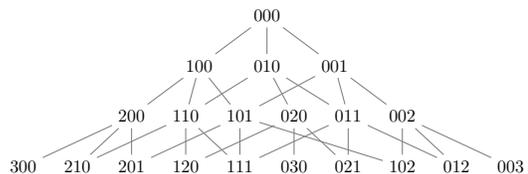

\begin{definition}[Pareto dominates]
  Let $t_1=(a_1,\dots, a_n)$
  and $t_2=(b_1,\dots, b_n)$
  be $n$-tuples of integers.
  We say that $t_1$ \emph{Pareto dominates}  $t_2$,
  if and only if  $ t_1\neq t_2$
   and $a_i\leq b_i$ for all $ i\in 1..n$.
\end{definition}




We focus  on traversals of the graph of
tuples where traversing an edge  increases  one of the indices.  Hence,
there is an edge from tuple $t_1$ to tuple $t_2$ iff $t_2$ is obtained by
increasing  either of the digits of $t_1$ by $1$;  see \autoref{fig:graph}.
This graph anchors our initial motivation that the order on the terms
pertaining to a single variable represents preference. Indeed, following down any
edge in this graph means going to a less preferred tuple.  We call this graph
the \emph{Pareto graph}.

So what does differentiate one traversal from another?  In
graph theory vernacular, a traversal is broad or deep.  In our
context a broad  traversal is more \emph{fair}  since it
alters terms of different variables  evenly.  A deep traversal is more  \emph{adventurous}
since it opts for less preferred, i.e., riskier, instantiations.

Fair strategies observe the Pareto ordering, meaning that no tuple dominates any
of the previous tuples.  For instance, the sequence
$(0, 0), (1, 0), (1, 0), (1, 1)$
 respects Pareto ordering but
$(0, 0), (0, 1), (1, 1), (1, 0)$
does not because  $(1,  0)$  Pareto-dominates $(1,1)$.
Note that both of these examples respect the Pareto graph in the sense that a
node is visited only if at least one of its predecessors has been visited.

In the remainder of the section we introduce  techniques considered in the experimental  evaluation in \autoref{sec:experiments}.
 On a technical note, in practice the number of possible candidates  per variable may vary,  but  for
the sake of clarity, we assume that  each variable has the same number of possible candidate terms.
This means that every element of the tuple (digit) is in the range $0..M$  for some fixed $M\in\naturals $.
Effectively, this means that we are  looking for systematic enumerations of tuples from the space $[0..M]^n$,
with a fixed set of $n$ variables.

\subsection{Stages by maximal digit~\cite{DBLP:conf/tacas/ReynoldsBF18}}%
\label{sub:maximal_digit}
A straightforward enumeration would be to  interpret  each $n$-tuple as an  $n$-digit
number and enumerate from $( 0,\dots , 0)$ to $(M,\dots , M)$ by increasing this
number by 1 at a time; this would yield  enumeration according to the
lexicographic order.  This is however highly unfair because  for large  values
of $M$  the most significant digits are changed very late.

Instead, we numerate numbers so that
the maximum possible digit increases in each stage. Within any given stage, the
tuples are ordered lexicographically.
For two variables the sequence begins as follows:
$(0, 0), (1, 0), (0, 1), (1, 1), (2, 0),\dots, (2, 2),\dots$.
This ordering observes the Pareto
domination and  the algorithm runs in constant space.

\subsection{Stages by sum of digits}%
\label{sub:sum_of_digits}

The maximum digit approach  mitigates unfairness in large value of $M$ (large number of candidate terms).
However, it still leads to imbalance with a large number of quantified
variables, i.e., with large tuples.  Indeed, even  with  $M=1$ already
10 variables  require $2^{10}$  iterations before the most significant digit is
changed.
The alternative is to  iterate over combinations stratified by the \emph{sum of
all the digits}. Effectively, this leads to a breadth first traversal of the Pareto
graph and its effect is  more pronounced with large number of variables.
The initial sequence is as follows
$(0,0,\dots,0),
 (1,0,\dots,0),
 (0,1,\dots,0),\dots$
 $(0,0,\dots,1),
 (2,0,\dots,0),
 (1,1,\dots,0),
 (0,2,\dots,0),\dots
 $.
  This ordering also observes the Pareto  domination  and can be calculated in constant space.


\subsection{Leximax}%
\label{sub:leximax}
Arguably the most fair strategy is  enumeration according to the  \emph{leximax order}~\cite{BARBARA198834}
since  all the variables are in equivalent roles:
let $t_1, t_2$ be $n$-tuples of integers.
We say that $t_1$  is \emph{leximax preferred} to $t_2$
if $t_1^{\downarrow}$ is lexicographically smaller than $t_2^\downarrow$,
where $t^{\downarrow}$ denotes $t$ sorted in a descending order.
Enumeration can be done in constant space since all permutations
of   any tuple are incomparable.
This enables us to stage the enumeration by  gradually worsening a sorted tuple
and enumerate lexicographically  all its permutations through standard means.
The  permutations are enumerated lexicographically.
So for two  variables the sequence starts as follows, $(0, 0),(0, 1),(1, 0),(1, 1),(0, 2),(2, 0)$.
Contrast that with  the sum of digits $(0, 0),(0, 1),(1, 0),(0, 2),(1, 1),(2, 0)$.


\subsection{Iterative Deepening  and Random-walk Search}%
\label{sub:iterative_deepening}
Strategies discussed so far  never violate  Pareto domination,
which would be  violated by depth-first  but that would have a large degree of unfairness.
Instead, we propose to use iterative deepening where the maximum depth is incremented by some
fixed parameter $k\in\naturals^+$.
Maximum depth $2$ yields
$(0,0),(0,1),(0,2),(1,1),(1,0),(2,0)$,
where $(1,0)$ Pareto-dominates $(1,1)$, even though it comes later in the sequence.
We also use random-walk traversal which is similar to DFS
but instead of a stack we use a set where the next element to be explored is chosen randomly.

\section{Discarding Redundant Instantiations}%
\label{sec:masking}

\begin{table*}[!t]

    \caption{\label{fig:table-eval}Summary of problems solved. Best non-portfolio results are in bold.}%

  \centering%
  \begin{tabular}{@{}crrrrrrrrrrrrrr@{}}
    \toprule
    Library &\multicolumn{1}{c}{\#}&\multicolumn{1}{c}{e}&\multicolumn{1}{c}{u}&\multicolumn{1}{c}{id2} &\multicolumn{1}{c}{id4}&\multicolumn{1}{c}{lmax}&\multicolumn{1}{c}{sum}&\multicolumn{1}{c}{rwlk}&&\multicolumn{1}{c}{allu-port}&\multicolumn{1}{l}{eu-port}&\multicolumn{1}{l}{eallu-port}&&\multicolumn{1}{c}{z3}\\
    \midrule
    TPTP  &18627&\textbf{7765}&6989&6801&6834&6832&6922&6839&&7330&9056&9292&&\multicolumn{1}{c}{-}\\
 \midrule
    UF    &7668 &\textbf{3243}&3016&2975&2963&2959&3009&2992&&3120&3433&3452&&2905\\
    UFLIA &10137&\textbf{7424}&6024&6018&5897&6001&5980&5994&&6188&7595&7615&&6912\\
    UFNIA &13509&5715&\textbf{7458}&7396&7384&7426&7437&7430&&7620&7740&7843&&6491\\
    \bottomrule
  \end{tabular}

\end{table*}

When solving input with quantified formulas,
SMT solvers are often hindered by an overabundance of generated instantiations.
Thus, it is highly important to avoid instantiations that are \emph{redundant}.
At a high level,
an instantiation is considered redundant if it does not help rule out models in the current context.
Methods for discovering redundant instantiations are particularly important in the context of enumerative instantiation,
where typically we are iterating over similar domains of terms on multiple instantiation rounds,
and are looking for the first instantiation that is not redundant.

In our implementation,
we consider three criteria for determining that an instantiation
$\varphi \cdot \{ x_1 \mapsto t_1, \ldots, x_n \mapsto t_n \}$ is redundant,
in increasing order of cost:
\begin{enumerate}
\item
(Duplicate Term Vector)
For each $\varphi$,
maintain a trie containing all term vectors of its previous instantiations.
If $( t_1, \ldots, t_n )$ is already in this trie, then
the instantiation is redundant.
\item
(Entailed)
As described in~\cite[Section 4.1]{DBLP:conf/tacas/ReynoldsBF18},
a fast incomplete method for entailment is used for discovering when an instantiation lemma
is already implied by the current set of constraints known by the SMT solver.
All instantiations that are entailed are considered redundant.
\item
(Duplicate Formula Modulo Rewriting)
Maintain a set of previous formulas returned by quantifier instantiation.
Construct the formula $\varphi \cdot \{ x_1 \mapsto t_1, \ldots, x_n \mapsto t_n \}$
and normalize it using rewriting techniques.
If the resulting formula is already in our set, it is redundant.
\end{enumerate}
If none of these criteria hold, the instantiation is not considered redundant.

It is important to note that the latter two methods allow one to learn that a \emph{class} of instantiations is redundant.
For this purpose, we introduce the concept of a \emph{fail mask} for an instantiation.
A fail mask $\mathcal{M}$ for a substitution $\{ x_1 \mapsto t_1, \ldots, x_n \mapsto t_n \}$
is a sequence of $n$ bits such that all substitutions that extend
$\{ x_i \mapsto t_i \mid \text{the }i^{th} \text{ bit of } \mathcal{M} \text{ is set } \}$
when applied to $\varphi$ result in a redundant instantiation.

For example, let $\varphi$ be the formula $P(x_1, x_2) \vee Q(x_2,x_3)$, and
consider the substitution $\sigma = \{ x_1 \mapsto a, x_2 \mapsto b, x_3 \mapsto
c \}$.  Let $E = \{ P(a,b), \neg Q(b,c) \}$ be the current set of assertions
from the ground solver.  The instantiation $\varphi \cdot \sigma$ is redundant;
a fail mask for $\sigma$ is $110$, since $P(a,b) \vee Q(b, x_3)$ is entailed by
$E$ for any value of $x_3$.

We incorporate fail masks into our implementation in the following way.
When an instantiation $\varphi \cdot \sigma$ is discovered to be redundant,
we construct the fail mask $\mathcal{M}$ containing all $1$s.
Starting with $i=1$,
we drop the entry $\{ x_i \mapsto t_i \}$ from $\sigma$.
If the instantiation is still redundant based on the latter two criteria
above, then we set the $i^{th}$ bit to $0$.
If not, then we re-add the entry $\{ x_i \mapsto t_i \}$ to $\sigma$,
and proceed with $i+1$.
Notice this means that our computation of the fail mask is greedy.

The fail mask is  incorporated into the enumerative  strategies as follows.
After each failed instantiation, combine the tuple of term indices  and the
fail mask into a tuple with wildcards, denoted  ``$?$''.  So for instance, if the
tuple $(5, 4, 3)$  fails with the mask $101$, construct the tuple $(5,?,3)$
meaning that if the first variable is instantiated with the $5^\text{th}$ term
and the third variable with the $3^\text{rd}$ term,  the instantiation is
bound to be redundant.  Such combinations we wish to  avoid.
 This is checked independently of the enumeration algorithm by storing the
disabled patterns into a trie and discarding any combinations matching  one
of the previously disabled patterns.  The trie handles the wildcard
character $?$ specially  by always matching on it.

\section{Experiments}%
\label{sec:experiments}

This section reports on our experimental evaluation of different tuple
enumeration strategies implemented in the cvc5 SMT solver (the successor of
CVC4~\cite{Barrett2011}).
We performed all experiments on a cluster with Intel Xeon CPU E5-2620 CPUs with
2.1GHz and 128GB memory, providing one core, 300 seconds, and 8GB RAM for each
job.

Benchmarks are selected from first-order benchmarks from the TPTP
library~\cite{Sutcliffe2009}, version 7.4.0, and from
SMT-LIB~\cite{Barrett2017}, 2020 release.
Of 19287 first-order TPTP problems, we excluded 660 which contained polymorphic
types, leaving 18627 for consideration.
For SMT-LIB, we considered all problems from logics containing quantifiers and
integer arithmetic, i.e., UF, UFLIA, and UFNIA, totaling 31314 problems.
This selection of benchmarks was inspired by the evaluation
from~\cite{DBLP:conf/tacas/ReynoldsBF18}, where enumerative instantiation was
shown more effective in the above sets.

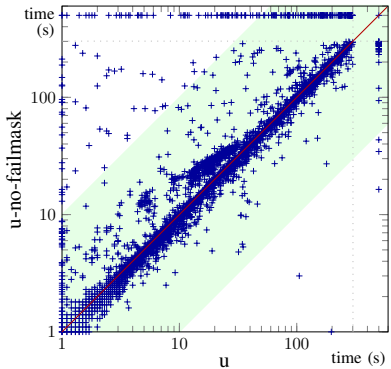
\begin{figure}[htbp]
  \centering

\definecolor{lightgreen}{rgb}{.9,1,.9}
\definecolor{darkblue}{rgb}{0,0,0.6}
\definecolor{darkred}{rgb}{0.7,0,0}
\begin{tikzpicture}[scale=0.8, every node/.style={scale=0.8}]

  \draw[color=lightgreen,fill=lightgreen] (0, 0) -- (1.95, 0) -- (5.43,3.48) -- (5.43,3.48) -- (5.43,5.43) -- (3.48,5.43) -- (0,1.95) -- (0,0);

  \begin{axis}[
    xmode=log,
    ymode=log,
    enlarge x limits=0,
    enlarge y limits=0,
    log ticks with fixed point,
    width=7cm,
    height=7cm,
    xmin=1,
    xmax=600,
    ymin=1,
    ymax=600,
]
    \addplot+[
    only marks,
    mark=+,
    darkblue,
    mark size=1.5pt] table {scatter.data};
  \end{axis}

  \draw[color=darkred] (0, 0) -- (5.4,5.4);

  \draw[color=lightgray,dotted] (0,4.835) -- (5.4,4.835);
  \draw[color=lightgray,dotted] (4.835,0) -- (4.835,5.4);

  \node at (2.7,-.5) {u};
  \node[rotate=90] at (-.8, 2.7) {u-no-failmask};

  \node[below left] at (5.5,-0.2) {{\footnotesize time (s)}};

  \node[left] at (0, 5.3) {{\footnotesize time}};
  \node[left] at (0, 5) {{\footnotesize (s)}};

\end{tikzpicture}

\caption{Impact of elimination of redundant instantiation via fail masks.}
  \label{fig:failmask}
\end{figure}

The evaluation covers a number of cvc5 configurations. The default enumeration, maximal digit,
 is denoted as \textbf{u}. Its
variations according to different enumeration strategies described above are
id-n for iterative deepening with increment~$n$; \textbf{lmax} for leximax;
\textbf{sum}  of digits; and \textbf{rwlk} for random walk.
We also run, for control, cvc5's E-matching (denoted \textbf{e}) and z3 4.8.10
(denoted \textbf{z3}).
All the cvc5 configurations use conflict-based
instantiation~\cite{Reynolds2014,Barbosa2017} as a ``fail-fast'' technique,
given its strong focusing effect.
The z3 evaluation is restricted to SMT-LIB, given its limited support for TPTP.

The results are summarized in \autoref{fig:table-eval}.
The column \textbf{allu-port} is a virtual best solver (\textbf{vbs}) of all the enumerative
configuration, \textbf{eu-port} of a vbs of only e and u, and eallu-port a vbs of all
cvc5 configurations.
We first emphasize the tremendous advantage in UFNIA of u over e, which can be
explained by many benchmarks needing instantiations with key arithmetic
constants, such as 0, to enable the necessary ground reasoning to solve the
problem.
However, a large number of these benchmarks may be impossible to solve via
E-matching alone: if matching needs to be done on terms containing arithmetic
operators, e.g.\ to match $x+1$ with $1$, E-matching will fail, whereas
enumerative instantiation would instantiate the formula regardless.
Moreover, the different enumeration strategies do lead to significant
orthogonality among the different configurations. The vbs of the enumerative
configurations versus u reduces the number of \emph{unsolved} problems in UFNIA
in almost 3\%, while eallu-port vs eu-port reduces the number of unsolved in
almost 2\%.
These improvements are also present in TPTP, with similar reductions in the
number of unsolved problems when considering all the enumeration strategies in a
virtual best solver.
This clearly shows the benefit of integrating into actual portfolios different
enumeration strategies rather than having just the default one.

We also evaluated an even more adventurous enumeration strategy than those in
\autoref{fig:table-eval}, which randomly changes the strategy at each
instantiation round, thus effectively simultaneously trying all the strategies.
This random strategy performs similarly to the others, but can be deeply
influenced by the random seed chosen for selecting a strategy each round, to the
extent that changing the seed from 0 to 7 makes it go, in UFLIA, from 6007
successes to 6047.
This further reinforces the usefulness of diversifying the set of strategies
used for quantifier instantiation in practice.



Discarding classes of redundant instantiations
using fail masks gives a clear advantage as illustrated in \autoref{fig:failmask} (default enumerative instantiation strategy, on all benchmarks).
Using the fail masks leads to 217 uniquely solved problems, whereas without it
only 31 problems are solved uniquely.
Moreover, a large number of commonly solved problems have very significant speed
ups, as the plot makes clear.
On problems where the fail masks do not help, the overhead of computing and
checking them is noticeable (see the often prevalent blue just below the red
line). However, it is far from a deterrent, given the significant gains.

\section{Conclusions}%
\label{sec:conclusions}

Enumerative instantiation is powerful, versatile, and offers a lot of freedom
for strategies.  We presented several ordering heuristics for instantiation that
contribute to the orthogonality of the strategies,
and ultimately improve the SMT solver's performance and robustness.
%
This is especially useful when
a user is willing to employ a barrage of solver configurations
to tackle a high-priority problem instance.
%

In future work, we plan to investigate the applications of enumerative instantiation strategies
for portfolio approaches to SMT solving.
We also would like to pursue more advanced techniques
where tuple and term orderings are not fixed,
and may be influenced by previous successes or failures.


\bibliographystyle{plain}
\bibliography{references}
\end{document}